\documentclass[journal]{IEEEtran}
\IEEEoverridecommandlockouts

\usepackage{comment}
\usepackage{caption}
\usepackage{stfloats} 
\usepackage{booktabs}
\usepackage{diagbox}
\usepackage{tabularx}

\usepackage{varwidth}
\usepackage{amsmath,amssymb,amsfonts,graphicx}

\usepackage{hyperref}
\hypersetup{colorlinks=true,citecolor = blue}

\usepackage{graphicx,epstopdf,caption,url}   

\usepackage{amssymb} 
\usepackage{array}    
\usepackage{arydshln} 
\usepackage{graphics}
\usepackage{color}   
\usepackage{graphicx}   
\usepackage{graphicx,epstopdf,caption,url}   
\usepackage{multirow}  

\begin{document}
%
\title{AI-Generated Content (AIGC): A Survey}

\author{
	Jiayang Wu, Wensheng Gan*, Zefeng Chen, Shicheng Wan, and Hong Lin \\
	
	\thanks{This research was supported in part by the National Natural Science Foundation of China (Nos. 62002136 and 62272196), Natural Science Foundation of Guangdong Province (No. 2022A1515011861), Fundamental Research Funds for the Central Universities of Jinan University (No. 21622416), the Young Scholar Program of Pazhou Lab (No. PZL2021KF0023), Engineering Research Center of Trustworthy AI, Ministry of Education (Jinan University), and Guangdong Key Laboratory for Data Security and Privacy Preserving.}

    \thanks{Jiayang Wu, Wensheng Gan, Zefeng Chen, and Hong Lin are with the College of Cyber Security, Jinan University, Guangzhou 510632, China; and also with Pazhou Lab, Guangzhou 510330, China.  (E-mail: wsgan001@gmail.com)} 
 
	\thanks{Shicheng Wan is with the School of Business Administration, South China University of Technology, Guangzhou 510641, China.}

	\thanks{Corresponding author: Wensheng Gan}
}

\maketitle

\begin{abstract}

To address the challenges of digital intelligence in the digital economy, artificial intelligence-generated content (AIGC) has emerged. AIGC uses artificial intelligence to assist or replace manual content generation by generating content based on user-inputted keywords or requirements. The development of large model algorithms has significantly strengthened the capabilities of AIGC, which makes AIGC products a promising generative tool and adds convenience to our lives. As an upstream technology, AIGC has unlimited potential to support different downstream applications. It is important to analyze AIGC's current capabilities and shortcomings to understand how it can be best utilized in future applications. Therefore, this paper provides an extensive overview of AIGC, covering its definition, essential conditions, cutting-edge capabilities, and advanced features. Moreover, it discusses the benefits of large-scale pre-trained models and the industrial chain of AIGC. Furthermore, the article explores the distinctions between auxiliary generation and automatic generation within AIGC, providing examples of text generation. The paper also examines the potential integration of AIGC with the Metaverse. Lastly, the article highlights existing issues and suggests some future directions for application.

\textit{Impact Statement--} It is necessary for academia and industry to take an overview of what AIGC is, how AIGC works, how AIGC changes our lifestyles, and what AIGC will be in the future. This article proposes a survey of AIGC from its definition, pros, cons, applications, current challenges, and future directions to answer these urgent questions. We summarize the existing major literature, which helps relevant researchers become familiar with and understand the existing works and unsolved problems. Based on the review of literature and the commercialization of scientific and research findings, we conduct some cutting-edge AIGC research. In particular, the challenges and future directions of AIGC can be helpful for developing AI. Relevant technologies of AIGC will boost the development of artificial intelligence, better serve human society, and achieve sustainable development.

\end{abstract}


\IEEEpeerreviewmaketitle

\section{Introduction}

With Web 3.0 still in its blooming stage \cite{webGan2023}, Artificial Intelligence (AI)\footnote{https://en.wikipedia.org/wiki/Artificial\_intelligence} has proven to be an effective tool for many challenging tasks, such as generating content, classification and understanding. In recent years, some advancements within AI have helped the technology complete more complex tasks than before, such as understanding input data and then generating content. Artificial Intelligence Generated Content (AIGC) is a new content creation method that complements traditional content creation approaches like Professional Generated Content (PGC) and User Generated Content (UGC) \cite{tu2021ugc,sun2022deep}. AIGC generates content according to AI technology to meet the requirements of users. It is supposed to be a promising technology with numerous applications. Understanding AIGC's capabilities and limitations, therefore, is critical to exploring its full potential.

\begin{figure}[ht]
    \centering	
    \includegraphics[clip,scale=0.31]{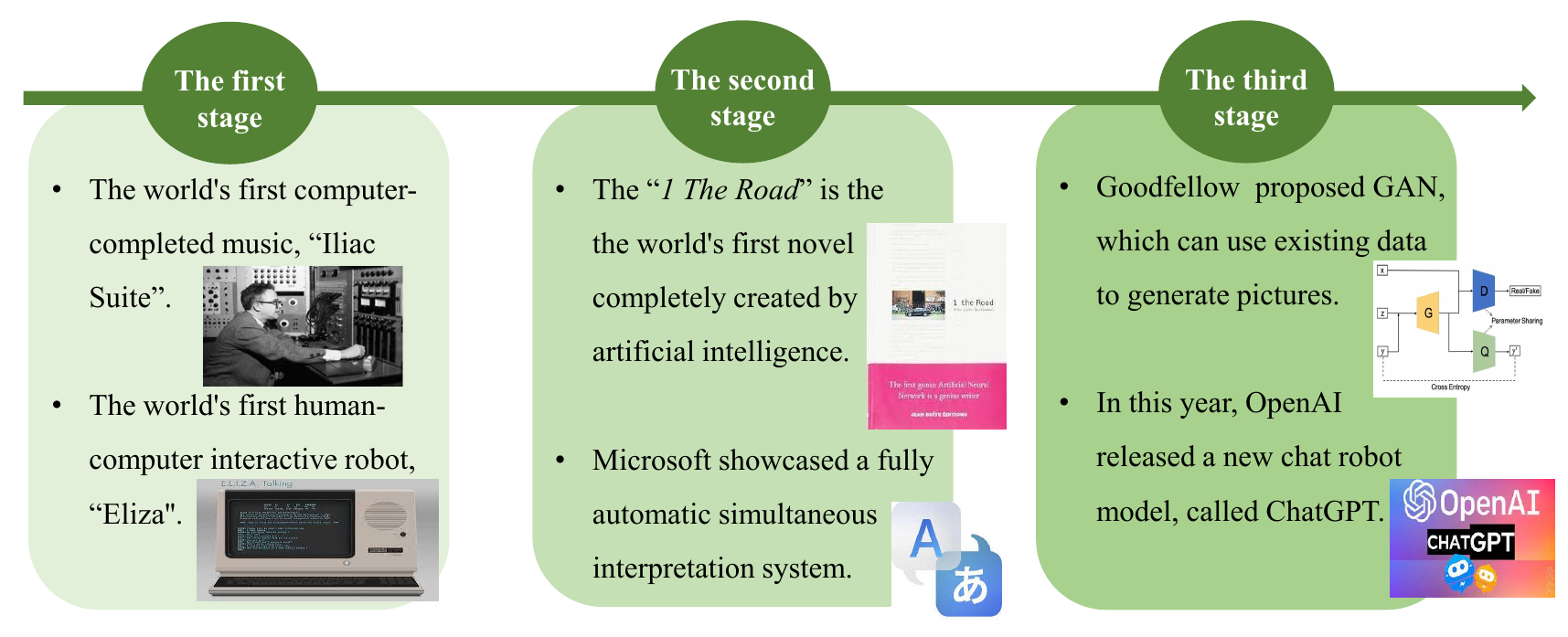}
    \caption{Three stages of AIGC.}
    \label{fig:aigc_history}
\end{figure}

Actually, the origins of AIGC can be traced back to an earlier time. The development history can be roughly divided into three stages (as shown in Fig. \ref{fig:aigc_history}). In the first stage, researchers control the computer to realize the output of content through the most primitive programming technology. Hiller and Isaacson completed the world's first computer-completed music, \textit{Iliac Suite}\footnote{\url{https://en.wikipedia.org/wiki/Illiac\_Suite}}, in 1957. Then, the world's first human-computer interactive robot, \textit{Eliza}\footnote{\url{https://en.wikipedia.org/wiki/ELIZA}}, came out. \textit{Eliza} shows the ability to search for appropriate answers through pattern matching and intelligent phrases but does not reflect a semantic understanding. However, most people still regard \textit{Eliza} as the sources of inspiration for AI nowadays. During the next two decades, it was the stage of sedimentation accumulation. The second stage assumes AIGC progress as usability as a result of the increased availability of massive databases and advancements in computing equipment performance. The Road\footnote{\url{https://en.wikipedia.org/wiki/1\_the\_Road}} is the world's first novel completely created by AI. After that, Microsoft also demonstrated a fully automatic simultaneous interpretation system, which is capable of translating speech from English to Chinese in a short time with high accuracy \cite{joshi2012cognitive}. However, the bottleneck of algorithms directly limits AIGC's ability to generate rich content. The third stage began in 2010 when AIGC entered a rapid development phase. Goodfellow \cite{goodfellow2020generative} proposed a Generic Adversarial Network (GAN), which uses existing data to generate pictures. In 2022, OpenAI released a new chat robot model, called ChatGPT. It is capable of understanding human language and generating text like humans do. Monthly active users exceeded 100 million within two months. There were about 13 million independent visitors using ChatGPT per day in January 2023\footnote{\url{https://www.theguardian.com/technology/2023/feb/02/chatgpt-100-million-users-open-ai-fastest-growing-app}}. With the improvement of products (like ChatGPT), AIGC has shown great potential for applications and commercial value. It has attracted a lot of attention from various domains, including entrepreneurs, investors, scholars, and the public.

At present, the quality of AIGC content is significantly better than it was before. Furthermore, the types of AIGC content have been enriched, including text, images, video, code, etc. Table \ref{table:light1} lists some AIGC models or classic products developed by major technology companies, as well as their corresponding applications. ChatGPT\footnote{\url{https://chat.openai.com/}} is a machine learning system based on the Large Language Model (LLM)\footnote{\url{https://en.wikipedia.org/wiki/Large\_Language\_Mode}}. After being trained on humorous large text datasets, LLM not only excels at generating reasonable dialogue but also produces compelling pieces (e.g., stories and articles). Thanks to its unique human feedback training process, ChatGPT is able to comprehend human thinking with greater precision. Google claims their upcoming product, Bard\footnote {\url{https://blog.google/technology/ai/bard-google-ai-search-updates/}} will have the same features but focus more on generating conversations. Compared to ChatGPT, Bard can make use of external knowledge sources, which can help users solve problems by providing answers to natural language questions instead of search results. In addition, Microsoft's Turning-NLG\footnote{\url{https://turing.microsoft.com/}} is an LLM with 17 billion parameters, and it is applied to summarization, translation, and question-answering.  

The Diffusion model is a cutting-edge method in the field of image generation. Its simplicity of interaction and fast generation features significantly lower the barriers to entry. Several popular applications, such as Disco Diffusion\footnote{\url{http://discodiffusion.com/}}, Stable Diffusion\footnote{\url{https://stablediffusionweb.com/}}, and Midjourney\footnote{\url{https://www.midjourney.com/}}, have generated exponential social media discussions and showcases of work. NVIDIA is a pioneer in visual generation research. Their product (i.e., StyleGAN) is a state-of-the-art approach to high-resolution image synthesis, specializing in image generation, art, and design. In addition, because of the distinct requirements for generating pictures within different industries, StyleGAN provides opportunities for several startups. For example, Looka focuses on logo and website design, and Lensa focuses on avatar generation. GAN is already capable of generating extremely realistic images. DeepMind is trying to apply it to the field of generating videos. Their proposed model, called Dual Video Discriminator GAN (DVD-GAN) \cite{clark2019adversarial}, can generate longer and higher resolution videos using computationally efficient discriminator decomposition. DVD-GAN is an exploration of realistic video generation.

\begin{table}[ht]
    \centering
    \caption{The AIGC and major technology companies}
    \label{table:light1}
    \begin{tabular}{|c|c|l|}
     \hline
     \textbf{Company} & \textbf{Product} & \qquad\qquad\textbf{Applications} \\
     \hline
     \hline
     \textbf{OpenAI} & ChatGPT & Text generation, chatbots, and \\
      &  &   text completion \\
     \hline
     \textbf{Google} & LaMDA & Question answering and chatbots \\
     \hline
     \textbf{NVIDIA} & StyleGAN & Image generation, art, and design \\
     \hline
     \textbf{Microsoft} & Turing-NLG & Summarization, translation, and \\ 
      &  &  question answering \\
     \hline
     \textbf{DeepMind} & DVD-GAN & Video generation \\
     \hline
     \textbf{Stability.AI} & Stable Diffusion & Text to images \\
     \hline
     \textbf{EleutherAI} & GPT-Neo & Text generation\\
     \hline
     \textbf{Baidu} & ERNIE & Question answering and chatbots \\
     \hline
    \end{tabular}
\end{table}

To provide more insights and ideas for related scholars and researchers, this survey focuses on the issues related to AIGC and summarizes the emerging concepts in this field. Furthermore, we discuss the potential challenges and problems that the future AIGC may meet, such as the lack of global consensus on ethical standards, and the potential risks of AI misuse and abuse. Finally, we propose promising directions for the development and deployment of AIGC. We suppose that AIGC will achieve more convenient services and a higher quality of life for humanity. The main contributions of this paper are as follows.

\begin{itemize}
    \item We present the definition of the AIGC and discuss its key conditions. We then illustrate three cutting-edge capabilities and six advanced features to show the great effect AIGC brings.

    \item We further describe the industrial chain of AIGC in detail and list several advantages of the large pre-trained models adopted in AIGC.
    
    \item To reveal the differences between auxiliary generation and automatic generation within AIGC, we provide an in-depth discussion and analysis of text generation, AI-assisted writing, and AI-generated writing examples.
    
    \item From the perspective of practical applications, we summarize the advantages and disadvantages of AIGC and then introduce the combination of AIGC and Metaverse.
    
    \item Finally, we highlight several problems that AIGC needs to solve at present and put forward some directions for future applications.
\end{itemize}

\begin{figure}[ht]
    \centering	
    \includegraphics[clip,scale=0.53]{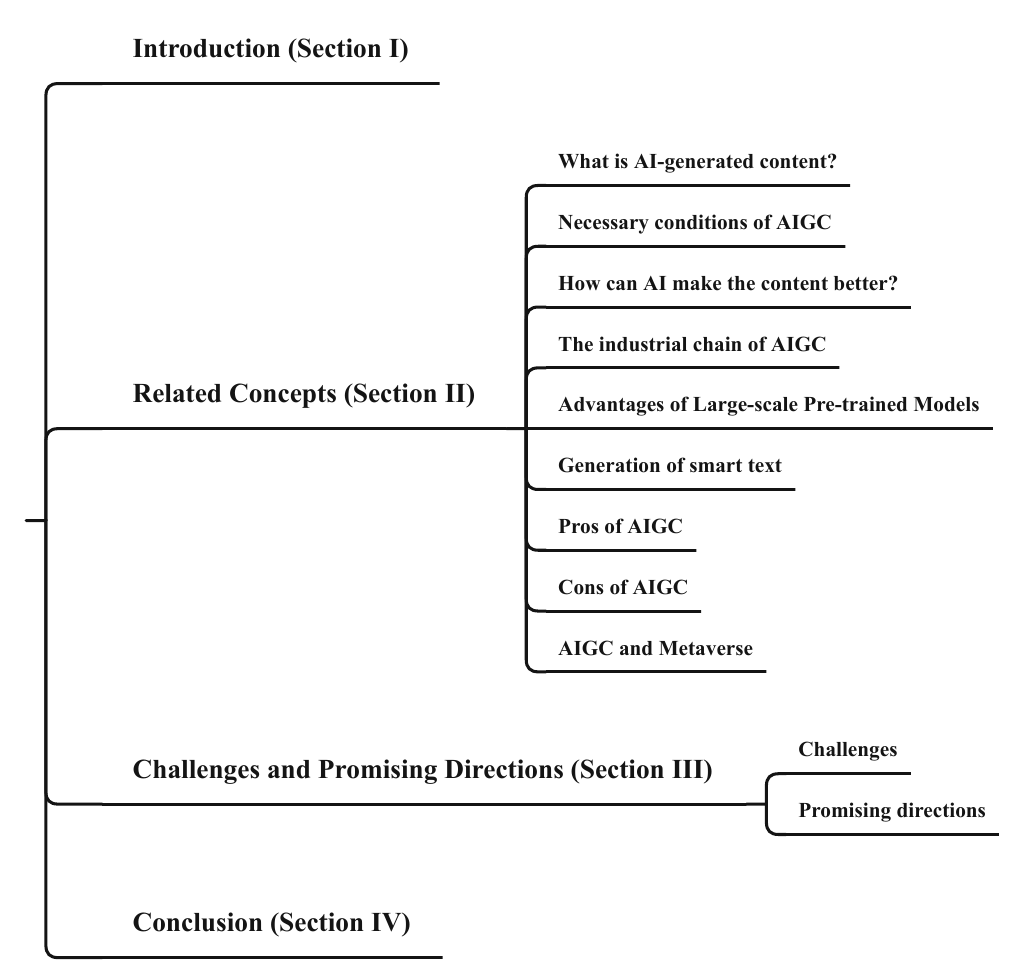}
    \caption{The outline of this survey.}
    \label{fig:outline}
\end{figure}

\textbf{Organization}: The rest of this article is organized as follows. In Section \ref{sec:relatedwork}, we discuss related concepts of the AIGC. We highlight the challenges in Section \ref{sec:challenges} and present several promising directions of the AIGC in Section \ref{sec:promisingdirections}. Finally, we conclude this paper in Section \ref{sec:conclusion}. The organization of this article is shown in Fig. \ref{fig:outline}.

\section{Related Concepts} \label{sec:relatedwork}

\subsection{What is AI-generated content?}

\begin{figure}[ht]
    \centering
    \includegraphics[clip,scale=0.42]{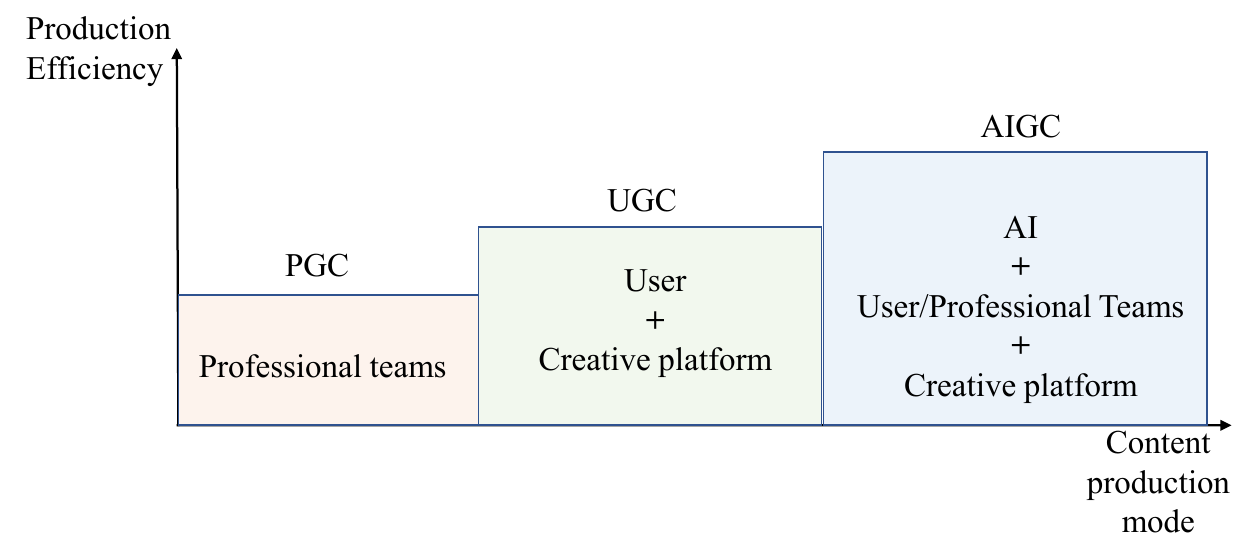}
    \caption{Three different modes of content production.}
    \label{fig:pgc_ugc_aigc}
\end{figure}

AI-generated content refers to writing pieces such as blogs, marketing materials, articles, and product descriptions that are created by machines. As shown in Fig. \ref{fig:pgc_ugc_aigc}, AIGC has experienced three different modes of content generation. In the PGC mode, content is generated by professional teams \cite{li2021pugcq,kim2012institutionalization}. The advantage of PGC is that most of the generated content is high quality, but the production cycle is long and difficult to meet the quantity demand for output. In the UGC mode, users can select many authoring tools to complete content generation by themselves \cite{gao2021user,welbourne2016science}. The advantage of UGC is that using these creative tools can reduce the threshold and cost of creation and improve the enthusiasm of users to participate in the creation. The disadvantage of UGC is that the quality of output content is difficult to ensure because the level of creators is uneven. AIGC can overcome the shortcomings of PGC and UGC in terms of quantity and quality. It is expected to become the primary mode of content generation in the future. In the AIGC mode, AI technology uses professional knowledge to improve the quality of content generation, which also saves time.

Some entrepreneurs are planning to use AIGC products to automatically finish advertisement production tasks, which was costly and time-consuming before. In general, AIGC can be categorized into text, picture, and video generation.

\textbf{Text generation.} AIGC encompasses structured writing, creative writing, and dialogue writing as its main subfields \cite{li2018seq2seq,see2017get,yu2017seqgan}. Structured writing primarily generates text content based on structured data for specific scenarios, such as news. However, creative writing involves generating text with a higher degree of openness, which demands personalization and creative capability. Creative writing is well-suited for marketing copy, social media, and blogs. Dialogue writing is mainly used for chatbots that interact with users through text. These bots are designed to answer questions, much like customer services.

\textbf{Pictures generation.} By leveraging AIGC, users can change and add new elements to their pictures based on the prompts given \cite{fan2020training,ren2021combiner,hoogeboom2021autoregressive}. It makes it easier and more efficient to edit images without the need for advanced skills or knowledge. Additionally, AIGC can independently generate images to meet specific requirements. For example, if a user needs a poster or logo in a specific format, AIGC can generate it in a short time. Another exciting application of AIGC is the creation of 3D models from 2D images \cite{pollefeys2002images}. 

\textbf{Audio generation.} AIGC's audio generation technology can be divided into two categories. That is text-to-speech synthesis and voice cloning, respectively. Text-to-speech synthesis needs input text and outputs the speech of a specific speaker. It is mainly used for robots and voice broadcasting tasks. Until now, text-to-speech tasks have been relatively mature. The quality of speech has met the natural standard. In the future, it will develop toward more emotional speech synthesis and small-sample speech learning. Voice cloning takes a given target speech as input, and then converts the input speech or text into the target speaker's speech. This type of task is used in intelligent dubbing and other similar scenarios to synthesize speech from a specific speaker. 

\textbf{Video generation.} AIGC has been utilized in video clip processing to generate trailers and promotional videos \cite{loeschcke2022text, texler2020interactive}. The workflow is similar to image generation, where each frame of the video is processed at the frame level, and then AI algorithms are utilized to detect video clips. AIGC's ability to generate engaging and highly effective promotional videos is enabled by the combination of different AI algorithms. With its advanced capabilities and growing popularity, AIGC is likely to continue to revolutionize the way video content is created and marketed.

\subsection{Necessary conditions of AIGC}

\begin{figure}[ht]
    \centering
    \includegraphics[clip,scale=0.31]{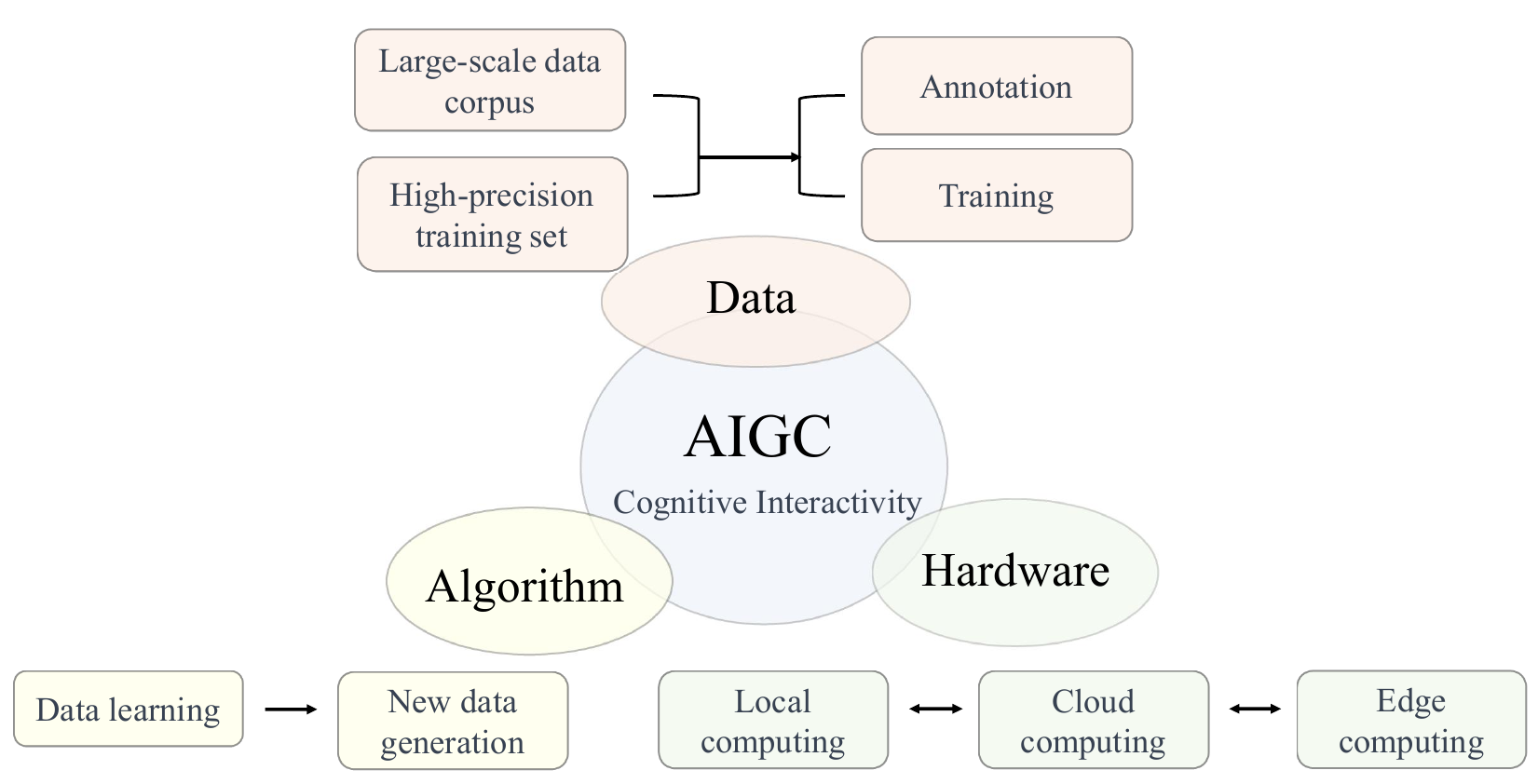}
    \caption{Relations between  hardware, algorithms, and data.}
    \label{fig:data_hardware_algorithm}
\end{figure}

As illustrated in Fig. \ref{fig:data_hardware_algorithm}, AIGC consists of three critical components: data, hardware, and algorithms. High-quality data, such as audio, text, and images, serve as the fundamental building blocks for training algorithms. The data volume and data sources have a vital impact on the accuracy of predictions \cite{zhang2010multi}. Hardware, particularly computing power, forms the infrastructure of AIGC. With the growing demand for computing power, faster and more powerful chips, as well as cloud computing solutions, have become essential. The hardware should be capable of processing terabytes of data and algorithms with millions of parameters. The combination of accelerating chips and cloud computing plays a vital role in providing the computing power required to efficiently run large models \cite{sze2017hardware}. Ultimately, the performance of algorithms determines the quality of content generation, and the support of data and hardware is crucial in achieving optimal results.

\textbf{Data.} The functionality of ChatGPT demonstrates that data is the foundation and basis for cloud computing and intelligent AI business iterations. The accuracy of training models depends on the size of the training datasets. The larger sample datasets often result in more accurate models. Typically, training tasks require billions to hundreds of billions of files. Therefore, storing and managing these massive datasets is crucial. To solve these issues, many cloud computing and data storage services, such as Amazon S3, Microsoft Azure Blob storage, and Google Cloud Storage, have been booming. Cloud storage services strongly offer storage solutions that are scalable, fast, secure, easy to process, and acceptable for massive data. Additionally, organizing and managing humorous datasets puts forward higher-level special techniques, such as data cleaning, duplicate data elimination, labeling, and categorization. All the above demands aim to make data well-organized and easier to process, thereby better supporting large-scale training and intelligent AI business applications.

\textbf{Hardware.} While massive data provides vital support for big data and AI applications, new storage demands are also urgent. The implementation of large models is heavily reliant on large computing power. Companies must consider the challenges of computing cost and algorithms' efficiency \cite{dally2018hardware}. Take ChatGPT as an example. ChatGPT can be divided into numerous AI models that require specific AI chips (e.g., GPU, FPGA, and ASIC) to handle complex computing tasks. According to OpenAI's report in 2020 \cite{brown2020language}, the total computational power required to train the GPT-3 XL model with 1.3 billion parameters is approximately 27.5 PFlop/s-day. Since ChatGPT is based on the fine-tuning of the GPT-3.5 model, which has a parameter quantity similar to the GPT-3 XL model. In other words, ChatGPT will take 27.5 days to complete the training at a speed of 1 trillion times per second. ChatGPT runs more than 30,000 Nvidia A100 GPUs to meet 13 million independent visitors per day in January 2023. The initial investment cost for these chips was approximate \$800 million, and the daily electricity charge is around \$50,000.

\textbf{Algorithm.} With the help of current intelligent data mining algorithms (e.g., neural networks \cite{bishop1994neural,lawrence1993introduction} and deep learning \cite{lecun2015deep,goodfellow2016deep}), the potential rules inherent in data can be learned independently by iteratively optimizing parameters within the learning paradigm and network structure. Moreover, with the development of the large-scale pre-training model, AI can combine the information from data mining to generate high-quality content. Large pre-training models are artificial intelligence models that use a large amount of text data for pre-training, such as BERT, GPT, etc. Large pre-training models are an important part of AI-generated content, and their improvement and development help to continuously improve the quality and accuracy of generated content. 

\begin{figure}[ht]
    \centering
    \includegraphics[clip,scale=0.32]{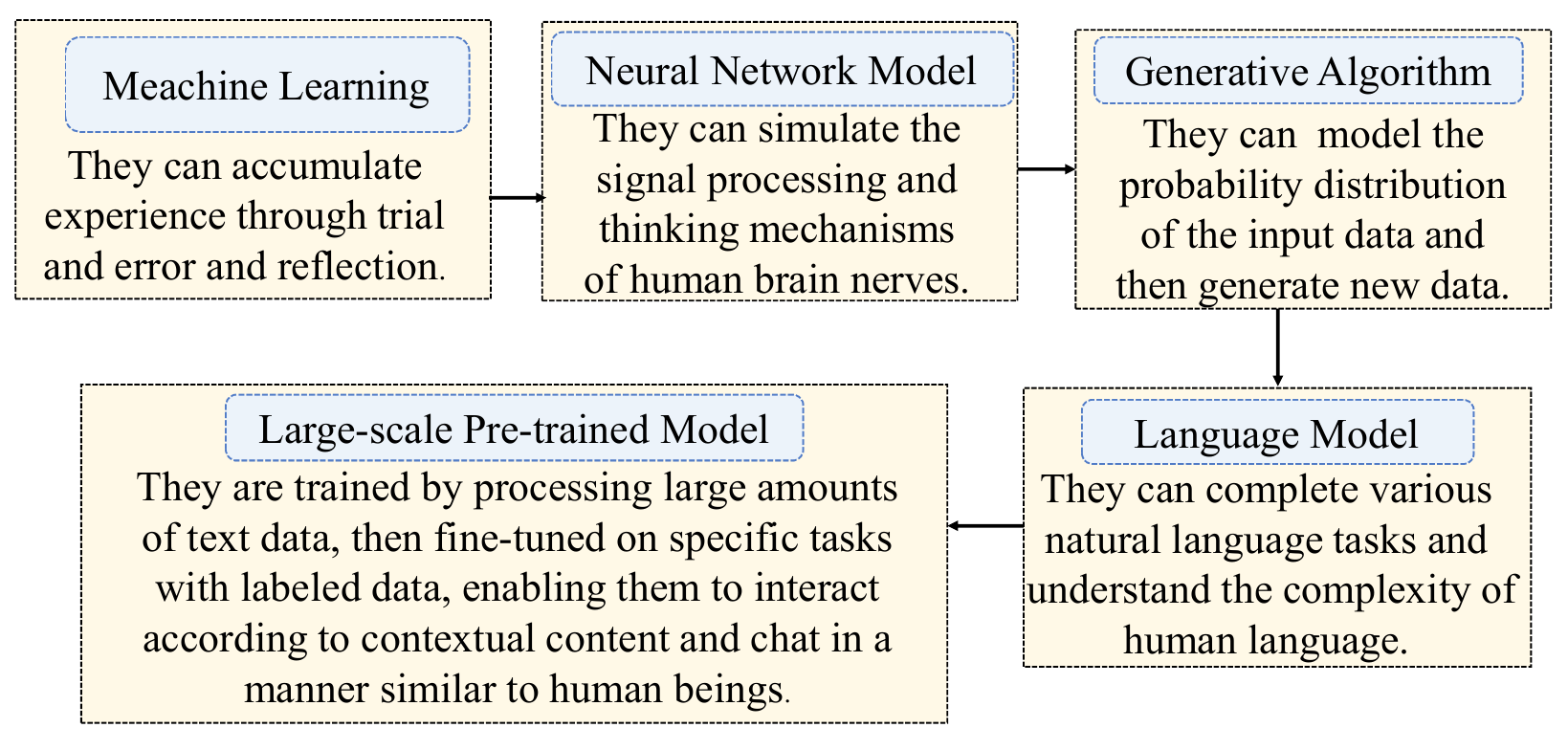}
    \caption{The evolution of AI models.}
    \label{fig:evolution_of_ai_algorithm}
\end{figure}

Actually, as shown in Fig. \ref{fig:evolution_of_ai_algorithm}, the current high-performance AI algorithm has gone through a long way of exploration. They gradually integrate the human thinking mode to improve the algorithm's efficiency. In traditional machine learning algorithms, data are classified by functions or parameters. The algorithms simulate the simple human brain, which improves the learning model through experience accumulation \cite{mitchell1997machine}. Neural network models further emulate the signal processing and thinking mechanisms of human brain nerves \cite{koutnik2014clockwork,chua1993cnn}. Furthermore, generative algorithms, such as Google's Transformer architecture \cite{kitaev2020reformer}, draw on human attention mechanisms to enable the completion of multiple tasks by an algorithm.

Goodfellow proposed the first generative model, Generative Adversarial Network (GAN), in 2014 \cite{goodfellow2020generative}. Table \ref{table:light} shows the evolution timeline of generative algorithms. In most cases, the significance of GAN is a source of inspiration for many popular variations and architectures. The transformer model has a wide range of applications in various domains (including NLP and CV). In addition, several pre-training models, such as BERT, GPT-3, and LaMDA, have been developed based on the Transformer model. The diffusion model is currently the most advanced image generation model because of its optimized performance.

With the development of generative models, language models have also made great progress. For example, Devlin \textit{et al.} \cite{devlin2018bert} proposed the BERT model to complete various natural language understanding tasks. BERT has revolutionary significance in understanding the complexity of human language. Furthermore, in recent years, there has been a rise in the popularity of large-scale pre-training models, which boast impressive generalization performance. Large-scale pre-training models can effectively address the challenges of frequent parameter modification. These models interact in a contextually-relevant manner and exhibit behavior similar to that of human beings when chatting and communicating because they are trained by connecting large-scale real corpora.

\begin{table*}[ht]
	\caption{The evolution timeline of generative algorithms.}
	\label{table:light}
	\begin{tabularx}{\textwidth}{m{2.5cm}<{\centering}|m{1cm}<{\centering}|m{13cm}<{\raggedright}}
		\toprule
		\hline
		\textbf{Algorithm} & \textbf{Year} & \multicolumn{1}{c}{\textbf{Description}}\\  
        \hline
        \textbf{VAE} \cite{kingma2013auto} & 2014 &  Encoder-Decoder models obtained based on variational lower bounds constraints. \\
        \hline
        \textbf{GAN} \cite{goodfellow2020generative} & 2014 & Generator-Discriminator models based on adversarial learning. \\
        \hline
        \textbf{Flow-based models} \cite{rezende2015variational} & 2015 & Learning a non-linear bijective transformation that maps training data to another space, where the distribution can be factorized. The entire model architecture relies on directly maximizing log likelihood to achieve this. The diffusion model has two processes, namely the forward diffusion process and the reverse diffusion process. During the forward diffusion phase, noise is gradually added to the image until it is completely corrupted into Gaussian noise. Then, during the reverse phase, the model learns the process of restoring the original image from Gaussian noise. After training, the model can use these denoising techniques to synthesize new ``clean'' data from random inputs. \\
        \hline
        \textbf{Diffusion} \cite{lehtinen2018noise2noise}  & 2015& The diffusion model has two processes. In the forward diffusion stage, noise is gradually applied to the image until the image is destroyed by complete Gaussian noise, and then in the reverse diffusion stage, the process of restoring the original image from Gaussian noise is learned. Following training, the model can use these denoising methods to generate new "clean" data from random input. \\
        \hline
        \textbf{Transformer} \cite{kitaev2020reformer} & 2017 & Originally used to complete text translation tasks between different languages, this neural network model is based on the self-attention mechanism. The main body includes the Encoder and Decoder parts, which are responsible for encoding the source language text and converting the encoding information into the target language text, respectively. \\
        \hline
        \textbf{Nerf} \cite{mildenhall2021nerf} & 2020 & It proposes a method to optimize the representation of a continuous 5D neural radiance field (volume density and view-dependent color at any continuous location) from a set of input images. The problem to be solved is how to generate images from new viewpoints, given a set of captured images. \\   
        \hline
        \textbf{CLIP} \cite{radford2021learning} & 2021 & Firstly, perform natural language understanding and computer vision analysis. Second, train the model with pre-labeled ''text-image'' training data. On the one hand, train the model on the text. From another aspect, train another model and continuously adjust the internal parameters of the two models so that the text and image feature values output by the models respectively match and confirm. \\
        \hline  \hline
        \end{tabularx}
\end{table*}

\subsection{How can AI make the content better?}

\begin{figure}[ht]
    \centering
    \includegraphics[clip,scale=0.3]{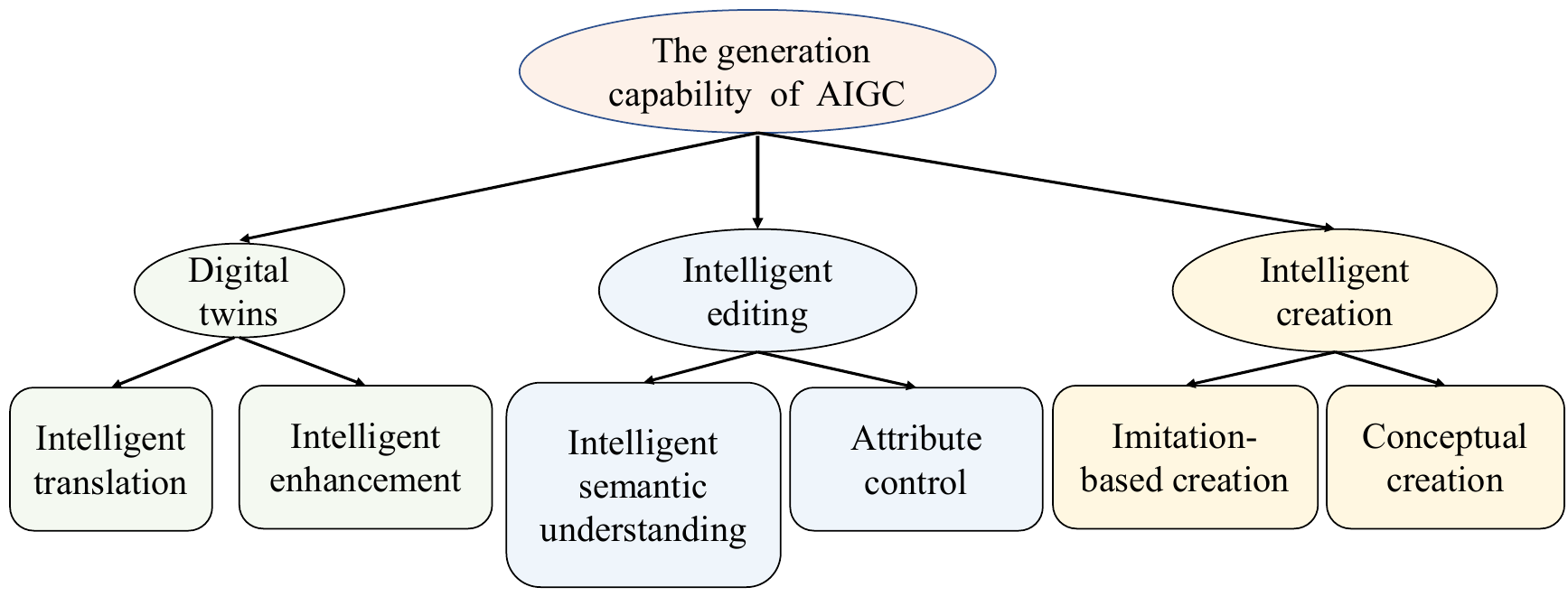}
    \caption{Three cutting-edge capabilities of AIGC.}
    \label{fig:three_cutting_edge_capabilities}
\end{figure}

AIGC owns three cutting-edge capabilities: digital twins, intelligent editing, and intelligent creation (Fig. \ref{fig:three_cutting_edge_capabilities}). These capabilities are nested and combined with each other to give AIGC superior generation capability.

\textbf{Digital twins.} AIGC can be used to map real-world content into the virtual world, such as intelligent translation and enhancement \cite{lv2022artificial,alexopoulos2020digital,el2018digital}. Intelligent translation involves transforming content across different modalities, e.g., language, audio, and visual, based on an understanding of the underlying meaning. This enables effective communication between people who speak different languages. Intelligent enhancement involves improving the quality and completeness of digitized content by filling in missing information, enhancing the image and audio quality, and removing noise and distortions. It is particularly effective when dealing with old or damaged content that may be incomplete or poor quality.

\textbf{Intelligent editing.} AIGC enables interaction between virtual and reality through intelligent semantic understanding and attribute control \cite{dong2016language,rabinovich2017abstract,yin2017syntactic}. Intelligent semantic understanding enables the separation and decoupling of digital content based on understanding. The attribute control enables precise modification and attribute editing based on understanding. The generated content can then be output into the real world, resulting in a closed loop of twinning and feedback. 

\textbf{Intelligent creation.} AIGC is a term used to describe the content generated by AI \cite{wu2008top,han2012data,olson2008advanced}. AIGC can be categorized into two types: imitation-based creation and conceptual creation. Imitation-based creation involves learning the patterns and data distribution features from existing examples. It creates new content based on previously learned patterns. Learning abstract concepts from massive data and applying studied knowledge to create new content that did not exist before is what conceptual creation entails. 

\begin{figure}[ht]
    \centering
    \includegraphics[clip,scale=0.38]{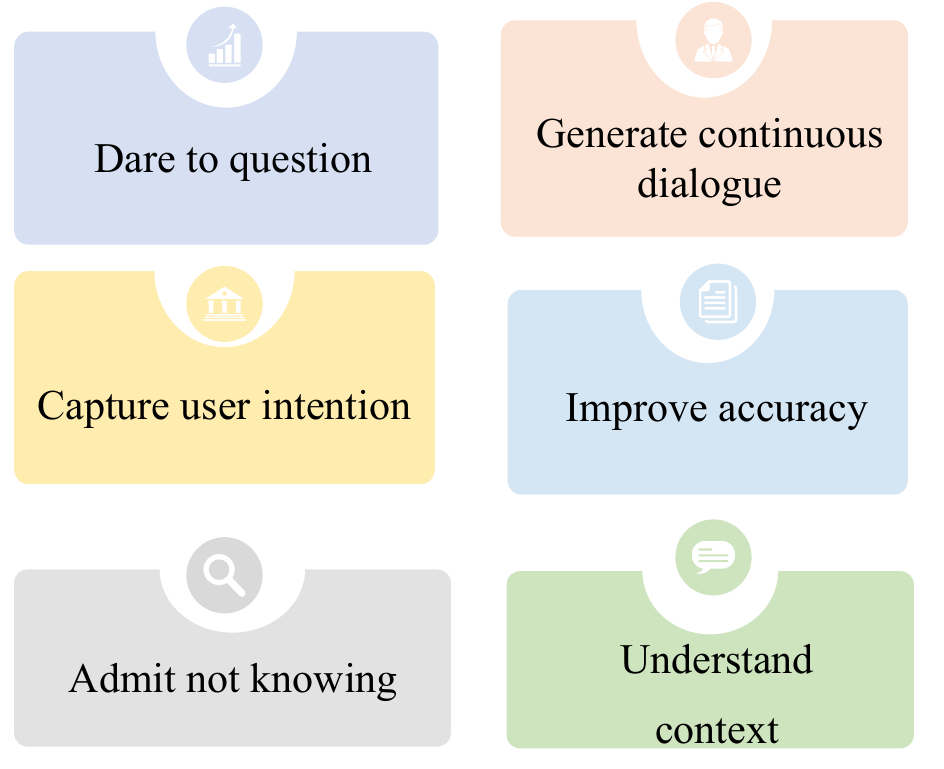}
    \caption{The six features of AIGC.}
    \label{fig:six_features}
\end{figure}

AIGC technology has become an increasingly popular tool for generating content in various industries. ChatGPT is an appropriate illustration of AIGC. Advanced reinforcement learning techniques and expert human supervision enable ChatGPT to acquire effective understanding and well-processed natural language. It was demonstrated to have a high degree of coherence in understanding the context. As shown in Fig. \ref{fig:six_features}, ChatGPT has six key features that make it a powerful tool in natural language processing. In terms of making conversations, ChatGPT can actively recall prior conversations to aid in answering hypothetical questions. Moreover, ChatGPT filters out sensitive information and provides recommendations for unanswered queries, which improves its usage performance. ChatGPT is an ideal tool for customer service, language translation, content creation, and other applications due to its advanced features. 

\subsection{The industrial chain of AIGC}

\begin{figure}[ht]
    \centering
    \includegraphics[clip,scale=0.31]{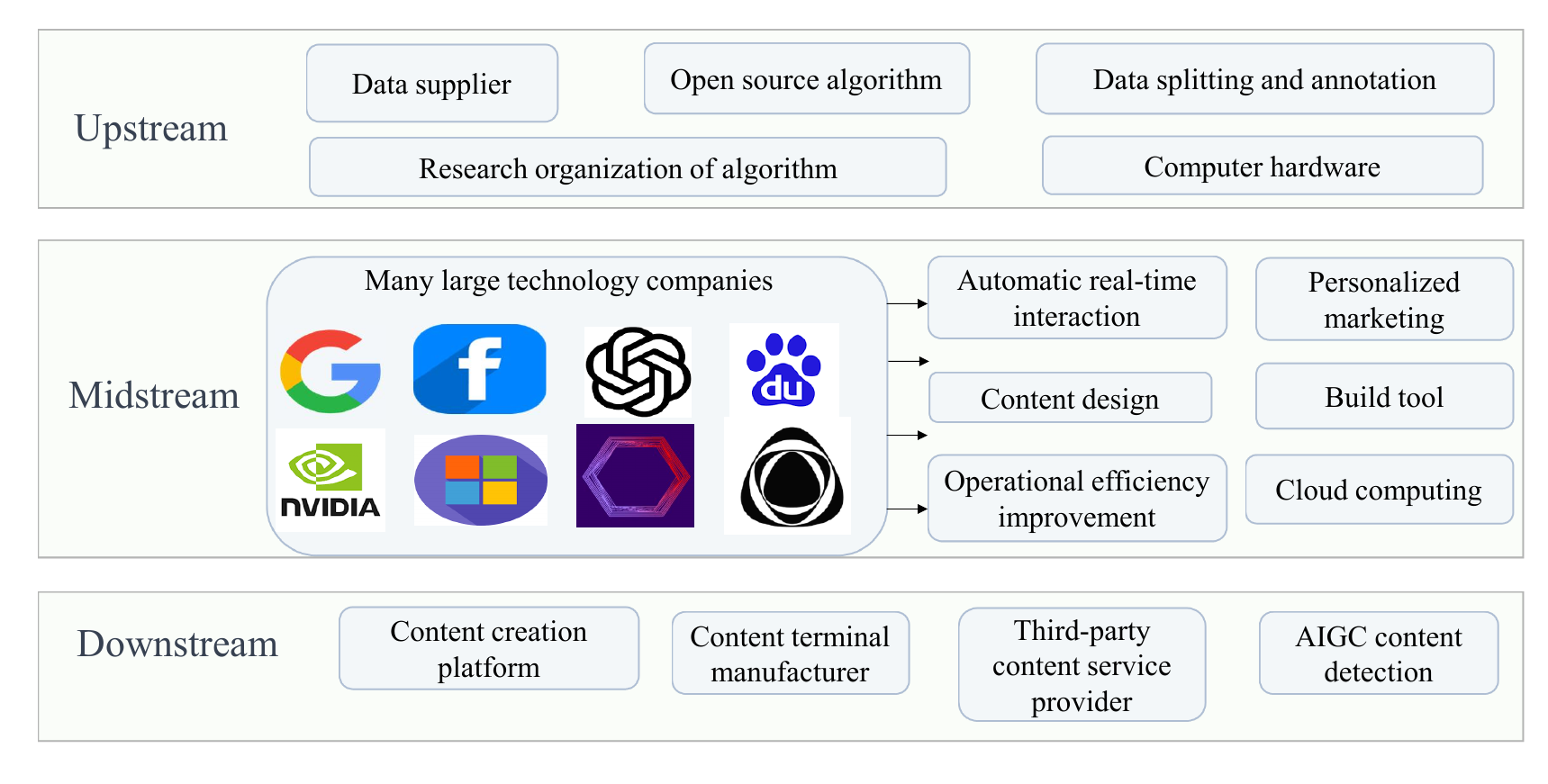}
    \caption{The industrial chain of AIGC.}
    \label{fig:aigc_chain}
\end{figure}

The AIGC industry chain is an interconnected ecosystem that spans from upstream to downstream. As shown in Fig. \ref{fig:aigc_chain}, downstream applications are heavily reliant on the basic support of upstream productions. Data suppliers, algorithmic institutions, and hardware development institutions are major parts of upstream AIGC. Data suppliers utilize web crawling technology to collect vast amounts of text from news websites, blogs, and social media \cite{shamrat2020effective}. Then, these wild data have to be automatically labeled or processed by NLP technologies \cite{sun2017review}. Algorithmic institutions typically consist of a group of experienced computer scientists and mathematicians with deep theoretical backgrounds and practical experience. They can develop efficient and accurate algorithms to solve various complex problems. Hardware development institutions focus on developing dedicated chips, processors, accelerator cards, and other hardware devices to accelerate the computing speed and response capabilities of AI algorithms.

The midstream sector includes big technology companies that integrate upstream data, hardware, and algorithms. These companies leverage these resources to deploy algorithms that set up computing resources and configure corresponding parameters in cloud computing, such as virtual machines, containers, databases, and storage. According to the specific properties and requirements of the algorithm, they ensure the optimal performance and efficiency of the algorithm through reasonable configuration. Then, the performance-optimized algorithm is encapsulated to generate a tool with an external interface. They are the bridge between upstream and downstream, connecting data suppliers and algorithmic institutions with content creation platforms and end-users. These companies earn revenue through personalized marketing, such as advertising placement and virtual brand building. In addition, midstream companies also play a critical role in advancing AI technologies. They invest in most research and development, which continuously enhances the performance and efficiency of AI systems. They also provide training data and feedback to upstream data suppliers and algorithmic institutions. The midstream companies contribute to the continuous improvement of the entire AIGC industry chain.

The downstream segment mainly consists of various content creation platforms. It lowers users' learning costs for creating content. Users can efficiently complete tasks with the help of midstream tools. For example, news media and financial institutions can quickly generate reports using text-generation tools. Since they are the primary recipients of the value that these technologies generate, downstream users are crucial in promoting the adoption and commercialization of AI technologies. By utilizing AI-powered tools and their services, downstream users can improve their productivity, enhance their decision-making, and create new opportunities for growth and innovation in their respective industries.

\subsection{Advantages of large-scale pre-trained models}

The large-scale AI model is a significant milestone in the development of AI towards general intelligence \cite{lin2022large}. The use of large-scale models is a clear indication of greater generalization for AIGC. Despite the challenges posed by the proliferation of general-purpose data and the lack of reliable data, deep learning entirely depends on models to automatically learn from data, and thus significantly improves performance \cite{l2017machine,ghahramani2015probabilistic}. Large-scale models possess both large-scale and pre-training characteristics and require pre-training on massive generalized data before modeling for practical tasks \cite{han2021pre}. These models are known as large-scale pre-trained models \cite{wang2023large}. In fact, AI's large-scale models can be seen as an emulation of the human brain, which is the source of AI's inspiration \cite{amit2018artificial}. In fact, the human brain is a large-scale model with basic cognitive abilities \cite{geary2021mitochondrial}. The human brain can efficiently process information from different senses and perform different cognitive tasks simultaneously. Thus, the AI large-scale model is not only expected to have numerous participants but also be able to effectively understand multimodal information, perceive across modalities, and migrate or execute between different tasks simultaneously. The improved accuracy of AI large-scale models in understanding human thinking is attributed to systems based on human feedback data for model training \cite{ma2020machine}.

\begin{figure*}[ht]
    \centering
    \includegraphics[clip,scale=0.3]{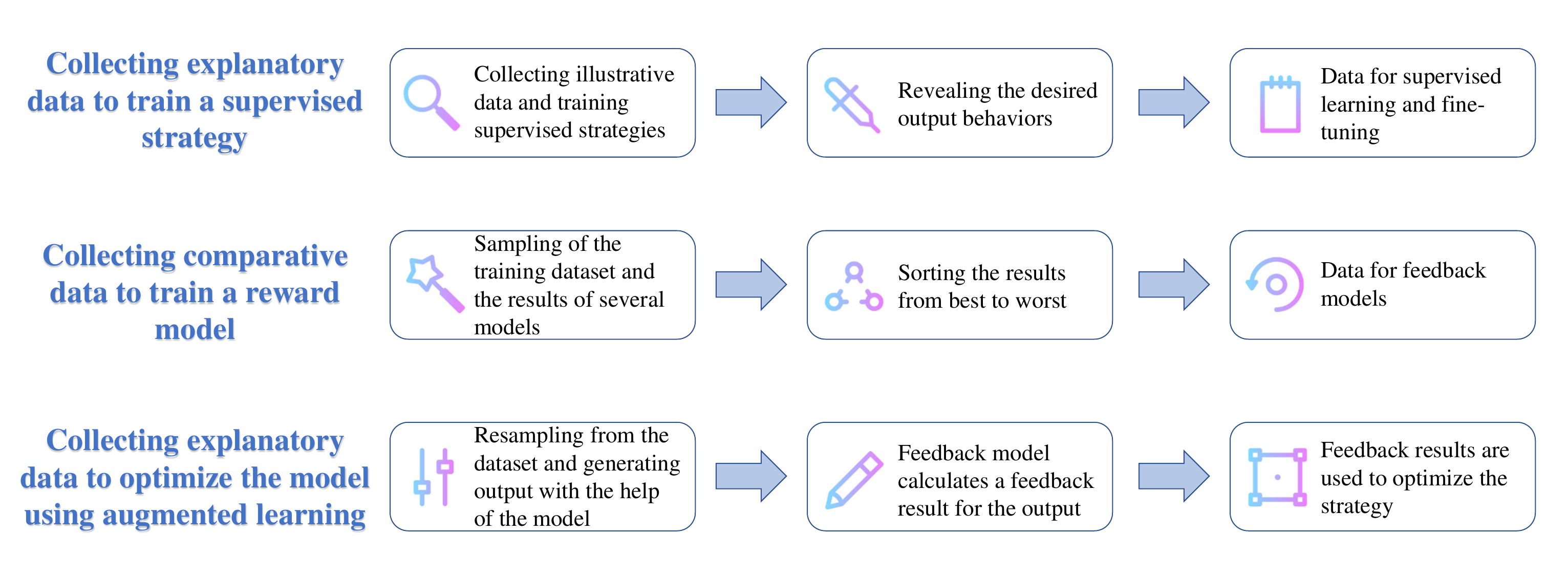}
    \caption{The specific steps of large-scale pre-trained models.}
    \label{fig:large}
\end{figure*}

As illustrated in Fig. \ref{fig:large}, the process of developing large-scale pre-trained models can be divided into three main steps. The first step is gathering explanatory data to train a supervised learning strategy. The second step involves collecting comparative data to train a reward model, which allows the model to make more accurate predictions. The final step is to collect explanatory data to optimize the model using augmented learning techniques. This will enhance the performance and efficiency of the model.

Hence, the use of large-scale pre-training models can improve the performance and generalization of AI. Specifically, large-scale pre-trained models have the following advantages for AI as well as AIGC:

\begin{itemize}
    \item  \textbf{Better generalization ability.} By pre-training on large-scale data, the model can learn more features and patterns, improving its generalization ability and allowing it to adapt to different tasks and scenarios.

    \item  \textbf{Save training cost.} The training cost of pre-trained models is relatively low because the data collection and labeling work only need to be performed once. The pre-trained models can be used for multiple tasks.

    \item  \textbf{Improve training efficiency.} Pre-trained models are fine-tuned in a fine-tuned way. Therefore, training can be done faster, and the results obtained can be better on smaller datasets.

    \item  \textbf{Support multiple tasks.} The pre-trained model can be used for multiple tasks, such as natural language processing, computer vision, and speech recognition. Due to fine-tuning training, these tasks greatly improve the applicability of the model.

    \item  \textbf{Continuous optimization.} The pre-trained model can be continuously optimized to expand the model's capability and make it more intelligent and adaptable by continuously adding new data and tasks.
\end{itemize}

\subsection{Generation of smart text} 

As previously mentioned, AIGC technology is unable to produce original content if you request specific needs and interests. Nevertheless, they can still play a useful role in the content creation process as writing assistants. We believe that there is a significant distinction between AI-assisted writing and AI-generated writing. 

\textbf{AI-assisted writing (AIAW).} The goal of AIAW is to provide assistance for human writing, which improves the coherence of the user's writing experience. This kind of writing tool can significantly improve the efficiency of writing in specific fields, such as legal documents. In fact, assisted writing can offer help in different stages of writing, like confirming the theme, writing content, and publishing the article. Before writing, the topic should be established first. The algorithm can recommend suitable text materials by analyzing current topics \cite{zhang2017hot}. This saves searching and sorting time. During the writing process, the algorithm can provide writing inspiration assistance \cite{jackson2015nature}. Through learning numerous similar articles, the AI model infers the subsequent parts of unfinished sentences from the perspective of statistical probability. AIGC can provide real-time error detection and correction suggestions for writing articles \cite{kim2019study} by collecting misspellings and incorrect word combinations from the corpus. Then, the algorithm provides comments on modifications to help authors improve their writing results. Before publishing, AIGC will add appropriate titles and labels about writing content.

\textbf{AI-generated writing (AIGW).} AIGW technology enables machines to write articles independently. Currently, computers are capable of automatically generating news alerts, hot press releases, and poetry articles. Intelligent writing algorithms can describe the main information contained in structured data. Due to the speed of machine processing, which is far faster than that of humans, AIGC works better in terms of generating time-sensitive news. For hot-draft writing, AIGC is useful in mining associated and related information \cite{zhang2015review}. AIGC can select appropriate content based on massive materials and extract relevant information through content analysis, ultimately organizing the results based on the writing logic \cite{zhang2015review}. Moreover, AIGC produces creative results that meet specific format requirements, including intelligent poetry writing and intelligent couplets \cite{he2012generating}. Because the model's output cannot be predicted in advance, AIGC has similar creativity to human writing. For example, if we want to use AIGC to write ancient poetry, we should input sufficient training data of poetry to train the model.

\begin{figure}[ht]
    \centering
    \includegraphics[clip,scale=0.36]{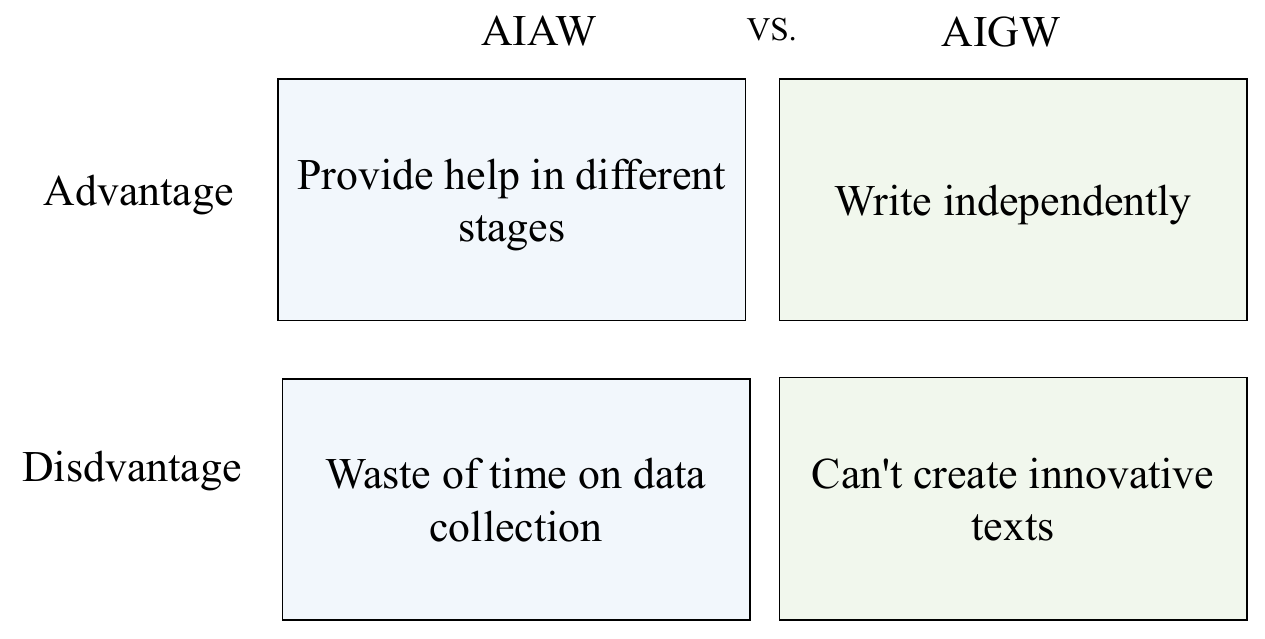}
    \caption{The differences between AIAW and AIGW.}
    \label{fig:aiaw_vs_aigw}
\end{figure}

\textbf{AIAW vs. AIGW.} The major differences between AIAW and AIGW are listed in Fig. \ref{fig:aiaw_vs_aigw}. Human beings have irreplaceable advantages in the field of writing. Deep learning models can easily create high-quality texts, but they cannot replace the subjective role of humans in writing practice. AI is superior to a human in data collection, but it cannot really create innovative, compassionate, and humorous texts. In addition, human writers have profound analytical abilities. Good writers not only have literary talent but also know how to effectively use words to express their thoughts. Human writers can purposefully decompose complex topics into easy-to-understand languages and provide valuable information for readers. Therefore, since AI tools are a valuable resource in content creation, it is important to balance usage, human creativity, and expression in writing. A reasonable division of labor between humans and machines is essential for achieving optimal results. In the future, AI should focus on data collection, while humans should be responsible for the creative process of writing.

\subsection{Pros of AIGC}

There are some pros to AIGC as shown in Table \ref{table:pros_of_aigc}. AI-generated content is becoming increasingly popular due to its strong abilities. AIGC is efficient, cost-effective, and even frees up human resources for other tasks. In this section, we will discuss some major benefits of AIGC.

\begin{table*}[ht]
    \caption{Several main pros of AIGC}
    \label{table:pros_of_aigc}
    \begin{tabularx}{\textwidth}{m{3.8cm}<{\centering}|m{13.5cm}<{\raggedright}}
    \hline
    \textbf{Pros} & \multicolumn{1}{c}{\textbf{Description}} \\
     \hline
     \hline
     \textbf{Efficiency and scalability} & AIGC can provide many benefits over traditional human writing, including speed and language localization. Another benefit of AIGC is its ability to create personalized social media posts for various sites.  \\
     \hline
     \textbf{Help scientific research} & AI can assist in analyzing large datasets through machine learning algorithms to identify patterns and correlations that might not be easily visible to humans.  \\
     \hline
     \textbf{For search engine optimization} & AI can analyze the content on a website and suggest changes to make it more SEO-friendly. \\
     \hline
     \textbf{Overcome writer's block} & AI tools can create detailed outlines and key points to help the writer determine what should be included in the article.  \\
     \hline
    \end{tabularx}
\end{table*}

\textbf{Efficiency and scalability}. AIGC can provide many benefits over traditional human writing, including speed and language localization \cite{adams2022artificial}. An AIGC production can produce an article in minutes, whereas a human writer will take a much longer time to finish it. This advantage allows AI tools to produce massive content in a short time. Additionally, AIGC can function in language localization according to translate content into a common language, which will be tailored to certain geographic areas. Another benefit of AIGC is its personalized social media creation ability. It is very useful for various websites. By analyzing users' online data, AI can create individual content for different users.

\textbf{Help scientific research.}  AIGC can have a significant impact on scientific research in multiple ways \cite{xu2021artificial}. Firstly, AI can assist in analyzing large datasets through machine learning algorithms to identify patterns and correlations that might not be easily visible to human researchers. Secondly, AI can analyze existing scientific literature and generate hypotheses that can be tested in further research, which can help identify new avenues for research. Additionally, scientists can use AI's learning ability in specific fields to get some research that benefits mankind. For example, AI can help in the development of new drugs and treatments by predicting the interactions between molecules and proteins. Overall, the use of AI-generated content can lead to more accurate and efficient research outcomes, saving time and resources in the process.

\textbf{For search engine optimization.} AI improves search engine optimization (SEO) in several ways \cite{sharma2019brief}. Due to the ability to provide data-driven insights and automate the workflow of AI, website owners are able to focus on creating high-quality content and providing better services. For example, AI-powered tools can analyze search queries and suggest relevant keywords to users. These tools make identifying patterns and trends easier by identifying keywords. AI tools optimize the length, structure, and readability of content, as well as suggest relevant keywords, to make websites more SEO-friendly.

\textbf{Overcome writer's block.} AI may be a helpful tool for writers solving writer's block according to inspiration, assistance, and polishing \cite{ippolito2022creative}. For instance, AI tools generate suggestions based on inputting keywords or topics. The tools analyze search data, trending topics, and popular queries to create fresh content. What's more, AIGC assists in writing articles and posting blogs on specific topics. While these tools may not be able to produce high-quality content by themselves, they can provide a starting point for a writer struggling with writer's block.

\subsection{Cons of AIGC}

One of the main concerns among the public is the potential lack of creativity and human touch in AIGC. In addition, AIGC sometimes lacks a nuanced understanding of language and context, which may lead to inaccuracies and misinterpretations. There are also concerns about the ethics and legality of using AIGC, particularly when it results in issues such as copyright infringement and data privacy. In this section, we will discuss some of the disadvantages of AIGC (Table \ref{table:cons_of_aigc}).

\begin{table*}[ht]
    \caption{Some major cons of AIGC}
    \label{table:cons_of_aigc}
    \begin{tabularx}{\textwidth}{m{3.8cm}<{\centering}|m{13.5cm}<{\raggedright}}
    \hline
    \textbf{Cons} & \multicolumn{1}{c}{\textbf{Description}} \\
     \hline
     \hline
     \textbf{Ethics and trust} & Due to the lack of intended tone and personality, the generated answers may be filtered out.\\
     \hline
     \textbf{Exacerbate social imbalances} & Some people can use AI tools to complete the original tasks at various speeds, whereas others may need to spend a significant amount of time thinking and creating content.   \\
     \hline
     \textbf{Negative effects on education} & AIGC may lack the human touch and personalization that are necessary for effective learning.  \\
     \hline
     \textbf{Inadequate empathy} & For instance, AI-generated music might not have the same emotional depth and authenticity as music performed and composed by humans.  \\
     \hline
     \textbf{Human involved} & People still need to be involved and articles quality-checked.  \\
     \hline
     \textbf{Missing creativeness} & It is hard for AIGC to come up with new content with the latest, trending ideas and topics.  \\
     \hline
    \end{tabularx}
\end{table*}

\textbf{Ethics and trust.} AI relies on data and algorithms to generate content, which may result in a lack of intended tone and personality \cite{yuan2018personality}. While AI tools can effectively cover the black-and-white areas of a topic, they may struggle with the more subjective gray areas. Additionally, plagiarized events will occur frequently, since AI tools often pull information from the same sources and reword it. To ensure authoritative and informative content, proper human review and curation are needed, especially if the information is pulled from various sources. The content can be crafted to maintain the intended tone, flow, and context by adding a human touch, thus improving the user experience.

\textbf{Exacerbate social imbalances.}  AIGC has the potential to exacerbate social imbalances. As a result, those who have access to and can afford advanced AI tools and technologies may have an unfair advantage over those who do not or cannot afford them. Some people can use AI tools to complete the original tasks at multiple speeds, while those who do not use AI tools may need to spend a lot of time thinking and creating content. This could lead to a situation where a small group of people dominates the production of content, creating a concentration of power and influence that can exacerbate existing inequalities.

\textbf{Negative effects on education.}  There are some potential negative effects of relying solely on AI-generated content for education \cite{tao2019artificial}. AIGC, for example, may lack the human touch and personalization required for effective learning. The use of AIGC can create a dependency on technology and discourage critical thinking and problem-solving skills. Students may become too reliant on the information provided by AI-generated content and fail to develop their own analytical skills. Additionally, AIGC may transfer the basic knowledge to users if the underlying data used to train the AI algorithms is biased or flawed.  It may cause students to form a permanently wrong knowledge system.

\textbf{Inadequate empathy.} While AI-generated content can be efficient and cost-effective, it may lack the creativity, emotion, and nuance that humans can bring to their creations. In the AI tool's work, it just generates content based on the parameters and objectives by analyzing large amounts of data and patterns, but it cannot truly comprehend the underlying meaning or context of the content. For example, compared to the music generated by composers and performers, AI-generated music may lack emotional depth and authenticity.

\textbf{Human involved.} While AIGC can certainly save time and effort in most cases, it is important to note that human involvement is still crucial in ensuring the quality and accuracy of the content \cite{bader2019algorithmic}. AI tools have the ability to aggregate information from multiple sources, but they may lack the nuanced understanding of language that humans possess. Because of this, the output can have mistakes and inconsistencies that need to be fixed by a person. For example, AIGC product descriptions may mix up textures and colors because of the tool's limited understanding of adjective meanings. 

\textbf{Missing creativity.} AI tools rely heavily on existing data to generate content, which can limit their ability to create fresh and original ideas \cite{jennings2010developing}. While they assist in streamlining content creation and generating ideas, they do not have the ability to generate completely new concepts on their own. This means that AIGC may not always be innovative or up-to-date with the latest trends. In other words, it may cause missing creativity. They may be able to analyze rich data to understand what types of content are popular or engaging, but they may not fully understand the nuances of a particular audience or be able to create content that truly resonates with them.

\subsection{AIGC and Metaverse}

The Metaverse \cite{mystakidis2022metaverse,chen2022metaverse} builds a persistent multi-user environment that combines physical reality with digital virtuality. It is a multi-user virtual space that allows multiple users to express their individual creativity. People communicate and interact with others through digital objects in a virtual environment \cite{mystakidis2022metaverse}. AIGC, in our opinion, can round out the Metaverse's personalized services and make it more vivid and vital.

AIGC enables efficient content creation, meets increasing demands for interaction, and improves personalized experiences \cite{david2014interactive}. It can simulate the virtual human brain to generate content for the Metaverse, including intelligent NPCs, automated QA, dialogue systems, and digital humans \cite{sun2022metaverse}. The Metaverse's concentration on cutting-edge technologies and users' interaction data accumulation can further enhance AIGC's intelligence and content creation abilities. By launching standardized and low-code development tools, AIGC enables small and medium-sized studios and individual developers to produce richer interactive content in the Metaverse.

In the Metaverse, the immense amount of data is the basis of maintaining smooth execution. With the help of AIGC technology, AI replaces humans to solve the Metaverse's needs in terms of massive data. Synthetic data based on AIGC technology has seen significant development in the Internet domain \cite{sun2022big}. AIGC data can be particularly useful in creating various scenarios within the Metaverse. For instance, considering an example of constructing a school online, a vast amount of environmental data is required to ensure a highly simulated scenario. Such work volume is a tedious and expensive process, which involves significant labor costs and resource utilization. However, AIGC can assist in generating the required environmental data, thereby saving a lot of time and money. By leveraging this process, AIGC contributes to the Metaverse data generation, which promotes the development of related technology in turn.

\section{Challenges} \label{sec:challenges}

\subsection{Data}

Data is one of the keys to ensuring the accuracy of training algorithms. In order to make output results more effective, the training datasets should ensure data quality and fairness \cite{bertossi2020data}. If the data contains deviations and inaccuracies in information, it may lead to biased and inaccurate responses, especially in terms of sensitive topics such as race, gender, and politics. To address this issue, synthetic data can be used in training. In the past, using real-world data to train AI models posed various problems, such as high costs for data collection and labeling, difficulty in ensuring data quality and diversity, and challenges in protecting privacy. Synthetic data can effectively solve these issues by serving as a cost-effective substitute for real-world data in training, testing, and validating AI models \cite{bolon2013review,nikolenko2021synthetic}. Using synthetic data not only makes training AI models more efficient but also enables AI models to self-learn and evolve in a virtual simulation world constructed from synthetic data.

When training with data, it is important to adhere to legal and ethical standards. If data collected through web scraping is used in large-scale model training, it is important to ensure that the data does not violate copyright or other legal regulations. If it only uses the public dataset, it is usually not necessary to obtain the consent of the copyright owner, as these data are already considered part of the public domain. However, if copyrighted data is used, it is necessary to obtain the permission of the copyright owner or provide appropriate compensation according to local legal regulations.

\subsection{Hardware}

The large-scale pre-training model's hardware problems are mainly troubling in two aspects: insufficient computing power and high energy consumption. The insufficient computing power problem is due to the models becoming increasingly complex. The number of parameters and calculation complexity are increasing exponentially, but hardware performance is not keeping up. In practice, high-performance computing devices such as GPUs and TPUs are required for the training and inference of large-scale pre-training models. However, even with these dedicated chips, it is difficult to meet the training and inference needs of super-large-scale models. In the paper published in 2020 \cite{brown2020language}, researchers from OpenAI reported that the pre-training of their language model GPT-3, which has 175 billion parameters, required 3.2 million core hours on a supercomputer with 2,048 CPUs and 2,048 GPUs. The inference of GPT-3 required a cluster of 2,048 CPUs and 2,048 GPUs, and the cost of running the model for a day was estimated to be around \$4,000.

The high energy consumption issue mainly stems from training and inference. Firstly, for the training phase, a large amount of computing resources are required to complete the model's training. This involves numerous matrix operations and neural network backpropagation. Secondly, for inference phase, due to the large number of parameters and complex calculation processes in large-scale pre-training models, the energy consumption of the inference phase is also high. Optimizing the calculation process and algorithm is a feasible approach to solving the above problems \cite{le2018mixed}. Utilizing efficient computing devices and technologies (e.g., mixed-precision computing and distributed training) also is another practical way \cite{lin2017deep}. 

\subsection{Algorithm}

One of the most significant advantages of large pre-trained language models is their ability to perform information retrieval tasks. In the past, information retrieval tasks were completed using a search-first-then-read approach. Reviewing several relevant contextual documents from external corpora is the first step. Then, answers were predicted from these documents. Due to powerful memory and reasoning skills, large language models significantly improve traditional steps \cite{bonifacio2022inpars}. Despite the significant progress made by large language models in information retrieval tasks, there are still some limitations that need to be addressed. For starters, a lack of vocabulary has an impact on retrieval accuracy and completeness. Since these models may only understand the vocabulary in the training data, specialized terms or new vocabulary may not be accurately comprehended. Second, contextual limitations cause the model to miss some implicit meanings and even cause some logical relationships to fail. To enhance the information retrieval capabilities of large language models, it is necessary to explore better language representation methods \cite{yang2019xlnet,zhuang2021robustly}.

To better meet user needs and handle complex tasks, the models should continuously improve and optimize themselves. The use of user feedback is an important part of the optimization algorithm \cite{stumpf2007toward,shuvo2022home}. Large pre-trained models can collect user responses by engaging them in feedback loops and using this feedback to optimize the model. This process typically involves presenting the model's prediction results to the user and requesting feedback. Feedback can be direct. For example, users can choose an option to indicate whether the prediction result is correct. Feedback can also be free-form. For instance, users can write a text to describe their views on the prediction result. Once enough feedback data is collected, the model can analyze this feedback to determine how to adjust the model. This process typically uses natural language processing techniques and machine learning algorithms to automatically analyze and summarize user feedback and transform it into data that can be used to optimize the model. 

When it comes to algorithms that generate content using AI, they are likely to be vulnerable to malicious attacks \cite{guembe2022emerging}. These attacks can take many forms, such as generating fake data or tampering with the generated content. Attackers can manipulate the model's input and output to deceive it and generate misleading content, which can affect the model's results and performance. This can lead to serious consequences such as the spread of misleading information, social engineering attacks, and forgery of evidence, among others. To address these attacks, it must improve the model's robustness and security, employ adversarial training techniques and encryption technologies, and increase user security awareness and vigilance \cite{hayes2017generating}.

\subsection{Privacy protection issues}

While training large pre-trained models, an unavoidable issue is how to rightly use sensitive personally identifiable information such as names, phone numbers, and addresses. During pre-training, this sensitive information may reflect the model's weights and parameters, which could be leaked to attackers or unauthorized third parties. Additionally, these large pre-trained models may also be used as the base models for text classification, sentiment analysis, and image recognition tasks, which further increases the risk of privacy breaches.

Moreover, distributed computing techniques are typically used to distribute the data across multiple computing nodes to relieve operation pressure. During this process, if appropriate security measures such as data encryption, access control, and data de-identification are not taken, attackers may obtain data by monitoring network traffic and attacking computing nodes \cite{chen2012data}. Therefore, a series of privacy protection measures need to be taken to protect the sensitive data contained in large pre-trained models, including but not limited to data de-identification, model security, restricting data access, and accountability. At the same time, corresponding security measures such as data encryption, access control, and data de-identification should be taken to maximize privacy protection when using these models \cite{jain2016big}.

\subsection{NLP for General AIGC}

\begin{figure}[ht]
    \centering
    \includegraphics[clip,scale=0.32]{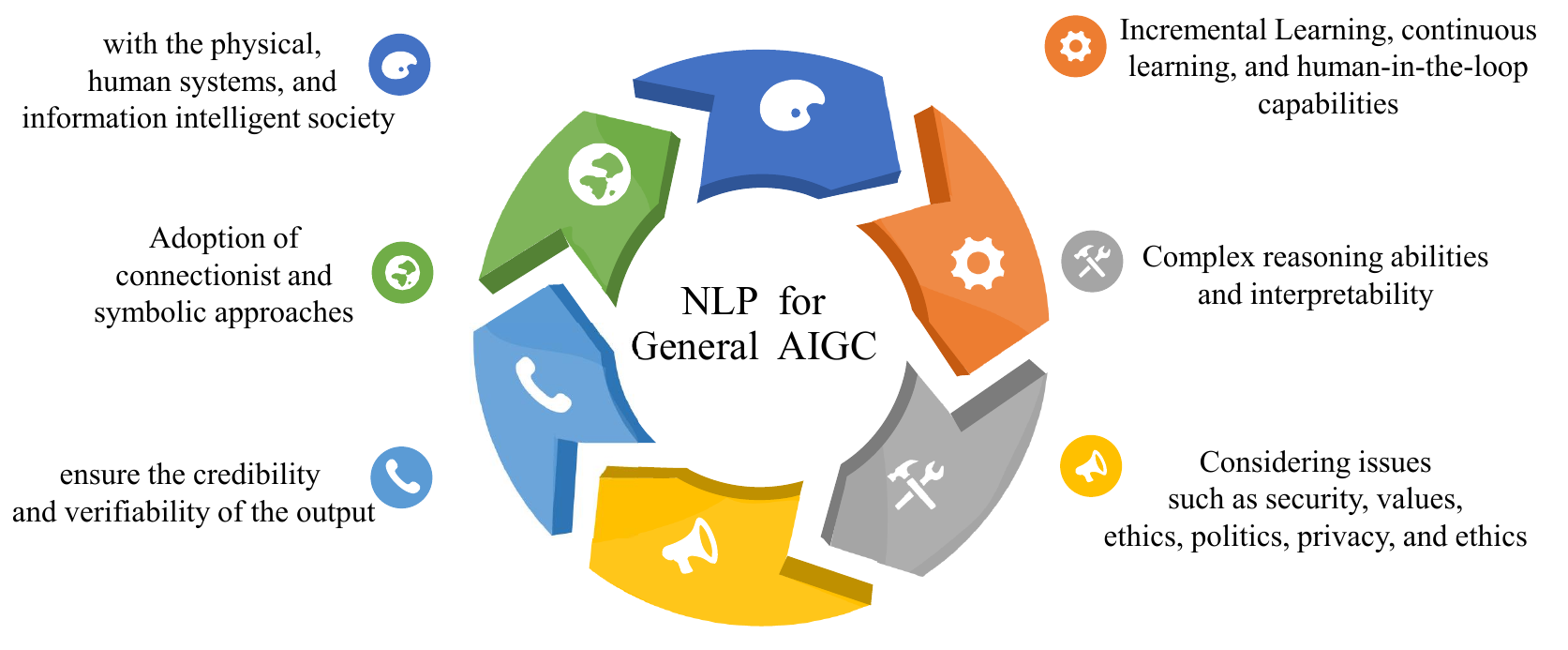}
    \caption{NLP for general AIGC.}
    \label{fig:general_aigc}
\end{figure}

With the continuous improvement of the capabilities of large language models \cite{chen2021evaluating}, natural language processing (NLP) faces many challenges (Figure \ref{fig:general_aigc}). In this era, we need a new generation of language models to further enhance the model's generation capabilities and then improve its descriptive ability and computability. Besides, carrying out a deep understanding of natural language (NLU) also needs the adoption of connectionist and symbolic approaches to solve various natural language processing tasks \cite{bates1995models}. On this basis, we need to ensure the credibility of the output results of NLP models, while also considering issues such as security, values, ethics, politics, privacy, and ethics.

To achieve these goals, it is essential to develop NLP models with complex reasoning abilities and interpretability. Addressing issues related to knowledge modeling, acquisition, and utilization can enhance the expressiveness and efficiency of these models. NLP models with incremental learning, continuous learning, and human-in-the-loop capabilities are also should be considered, as well as the creation of smaller models, model editing, domain adaptation, domain-specific models, and models tailored to particular applications and tasks. Furthermore, it is crucial to prioritize human-well learning and alignment to ensure the alignment of NLP technology with physical, human systems, and the intelligent society of information. By focusing on these aspects, we can make significant progress in advancing the field of NLP and ensuring it benefits society as a whole. In the era of large language models, the application of in-contextual learning (ICL) \cite{dong2022survey,sun2023does} has emerged as a new paradigm in natural language processing. By incorporating ICL into large language models, these models show a better understanding of context and produce more accurate and relevant results. Therefore, it is crucial to consider the use of ICL in improving the performance of NLP models.

\subsection{Human attitudes towards AIGC} 

\begin{figure}[ht]
    \centering
    \includegraphics[clip,scale=0.37]{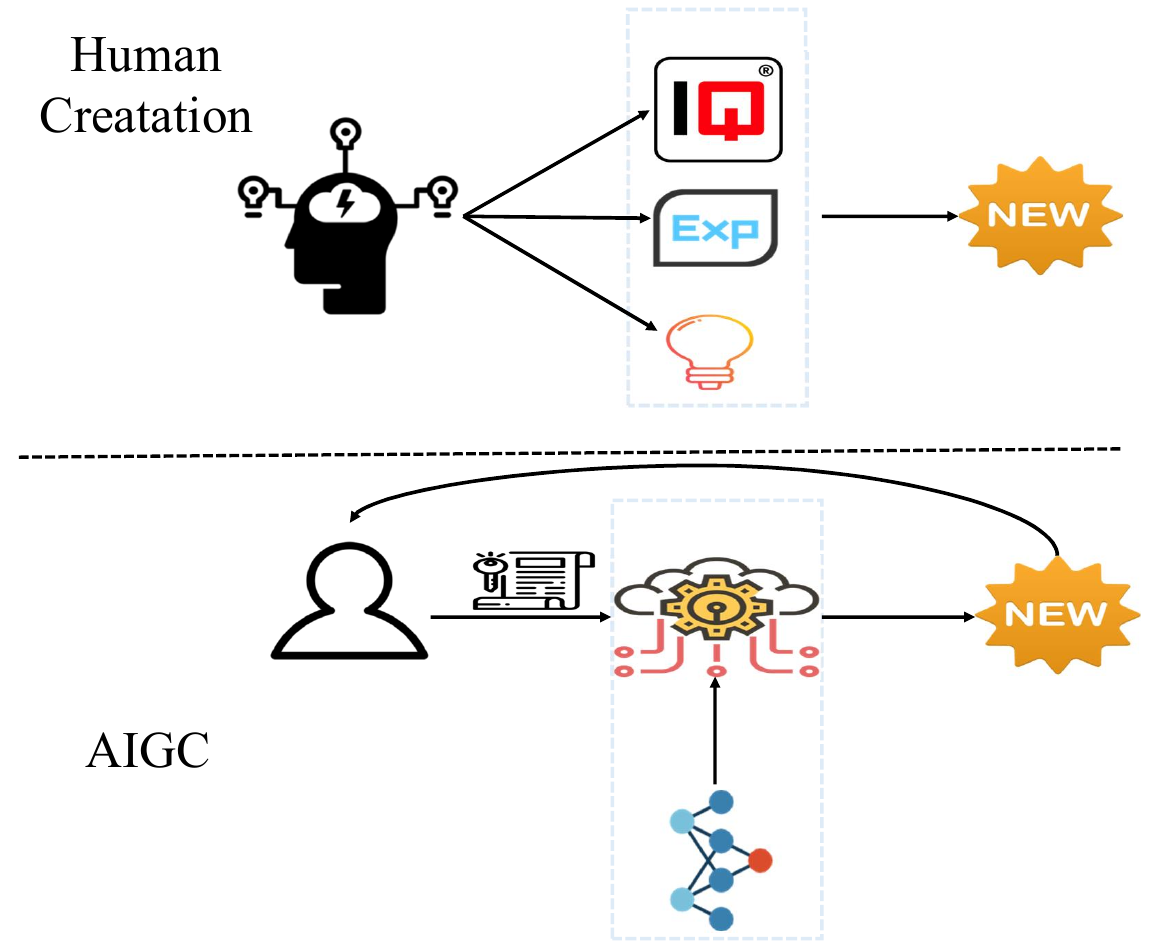}
    \caption{Human creation vs. AIGC.}
    \label{fig:human_creation_vs_aigc}
\end{figure}

There is a distinction to be made between AIGC and human-generated content, as illustrated in Fig. \ref{fig:human_creation_vs_aigc}. Human-generated content is the product of human intelligence, experience, creativity, and intuitive thinking. From another aspect, AIGC utilizes AI technology to train models to learn and simulate humorous data, analyze problems, and behave like humans. 

What aspects of AIGC need to be regulated by legislation? The first one is the ownership of creating content. At present, AIGC has taken the lead in media, e-commerce, film and television, entertainment, and other industries with high digitalization degrees and rich content demand to achieve significant development, and its market potential is gradually emerging. Using AIGC to automatically generate videos, music, and even computer games to make profits. Who does the income belong to? Is it the users or AI? Governments need to clarify protection rules about the intellectual property and data rights of AIGC which are based on the development and applications of AIGC technology. Since the commercial application of AIGC will mature quickly and the market scale will grow rapidly, the second aspect is that pursuing profit will cause people to spread rumors and make forgery easier than before. This urges governments to formulate appropriate policies (including positive and negative requirements). The policies should supervise programmers to take control and safety measures to ensure safe and controllable AIGC applications. More importantly, adopting content identification, content traceability, and other technologies to ensure a reliable source of AIGC is needed.

What is the scope of AIGC's activities? The key advantage of AI systems over other software systems is their superior efficiency. AI products have demonstrated the ability to perform tasks that are beyond the capacity of humans, such as creating hundreds of unique images within an hour or producing billions of words in a single morning. However, these capabilities have also raised concerns among many people. As we are all aware, technology is a double-edged sword that can either enhance human life or have detrimental consequences. Therefore, it is imperative to not only establish laws but also consider the morality of users when designing AI products. Several unethical incidents have occurred due to the use of AIGC products, including cheating, plagiarism, and discrimination. As a result, it is necessary to promote the ethical development of AI. Industry organizations can aid this effort by creating ethical guidelines for trustworthy AIGC. Additionally, programmers developing AIGC applications should follow the ``ethics by design'' paradigm. Finally, the ethics committee must establish a comprehensive and universal ethical review system.

What is the relationship between humans and AI? We are very much in the habit of seeing ourselves in the world around us. And while we are busy seeing ourselves by assigning human traits to things that are not, we risk being blindsided. As the saying goes, ``a coin has two sides''. The ChatGPT has jolted some people out of secure jobs and made them fearful of losing their jobs. At the same time, it makes AI employees see the dawn of AI. Since an AI-generated picture surprisingly beat other contestants \cite{roose2022ai}, some critics suppose AIGC will lower the creativity of humans. Until now, there has been a trade-off: you accept the disadvantages of AI products in order to get the benefits they bring. The powerful function of AIGC may make workers lazy and rest on their laurels. Furthermore, it will discourage enthusiasm among new blood in industries. However, we tend to regard different AIGC applications as strong assistants. The growth of AI-powered data-driven technologies will bring more opportunities for most people. The bloom of the car industry causes thousands of new jobs to be created, which is far more than that of raising horses. AI will be a powerful ally for humans if we establish a comprehensive AIGC governance system.

\subsection{Trusted AIGC}

Large language models can provide detailed and informative responses to various complex questions. However, surveys indicate that these models may generate inaccurate and biased answers due to some reasons \cite{shen2023chatgpt,austin2021program}. For example, poor-quality data may be collected, and thus the model may not be able to differentiate the credibility of information sources or even assign a higher weight to unreliable information sources. Moreover, errors may also occur because of training. The model cannot determine whether the generated answer complies with ethical standards. Unfortunately, current algorithms cannot effectively solve the above issues. Humans checking the final answers are still indispensable.

Recently, ChatGPT was used to summarize a systematic review of the effectiveness of cognitive-behavioral therapy (CBT)\footnote{\url{https://en.wikipedia.org/wiki/Cognitive\_behavioral\_therapy}} on anxiety-related diseases published in \textit{JAMA Psychiatry}. However, ChatGPT provided some responses that contained factual errors, false statements, and false data. For instance, ChatGPT erroneously stated that the review was based on 46 studies, where it was based on 69. In addition, it overstated the effectiveness of CBT, which could have serious consequences, such as misleading academic research and affecting medical diagnosis and treatment. Moreover, if ChatGPT generates unethical responses, it could affect people's values and have a significant negative impact on society, such as endangering social security when lawbreakers ask ChatGPT questions about retaliation and terrorist attacks. Therefore, filtering out harmful responses is essential in improving the algorithms/models.

In the future, it is important to improve the transparency of large language models \cite{wu2022ai,janssen2020data}. Currently, the training sets and large language models used by these algorithms are not publicly available. Technology companies may conceal the internal operations of their dialogic AI and generate answers that contradict reality. These practices run counter to the trend of transparency in open science. To address these issues, we propose that scientific research institutions, including scientific funding organizations, universities, non-governmental organizations, government research institutions, the United Nations, and technology companies should collaborate to develop advanced, open-source, transparent, and democratically controlled algorithm models. By doing so, we can ensure that these models are trustworthy, reliable, and accountable to the public, while also promoting openness and transparency in the AI domain.

The source code of open-source large models can be used for free by anyone, which means that the organizations need to be responsible for the code, as the users of these models may use them for various purposes, including commercial or malicious purposes. As contributors or maintainers, they should ensure that the code is stable, reliable, and secure to prevent any negative impact from improper use. To ensure responsibility for the code, the organizations should add an appropriate license that explicitly allows or prohibits certain use cases. They should also stay closely connected with the community to understand the usage of the code and promptly address any potential issues. Finally, they should always be vigilant against potential abuse and malicious behavior and take measures to prevent them.

\section{Promising Directions} \label{sec:promisingdirections}

\begin{figure}[ht]
    \centering
    \includegraphics[clip,scale=0.52]{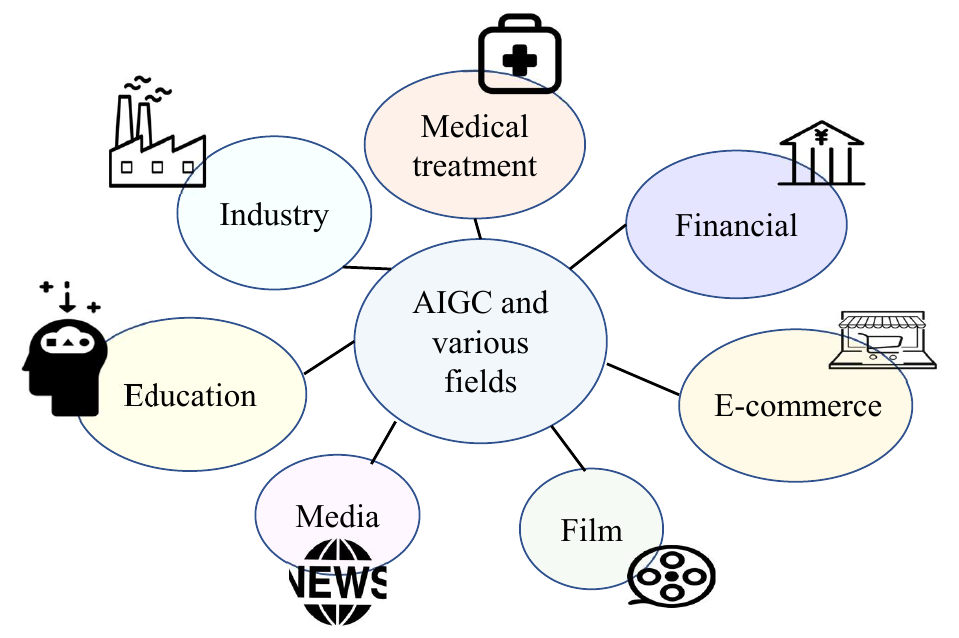}
    \caption{The combinations of AIGC and other fields.}
    \label{fig:aigc_and_various_fields}
\end{figure}

With the rapid development of hardware and algorithms, the future of AIGC is expected to see even more substantive applications. We believe that the most promising directions for AIGC include cross-modal generation, search engine optimization, media production, e-commerce, film production, and other fields, as illustrated in Fig. \ref{fig:aigc_and_various_fields}.

\subsection{Cross-modal generation technology}

The information present in the real world is a complex system comprising text, audio, vision, sensors, and human tactile senses. To accurately simulate the real world, it is necessary to utilize cross-modal generation capabilities. The development of large-scale pre-training models has enabled the maturation of cross-modal generation. Text-to-images and text-to-video are classic examples of cross-modal generation, which involve generating visual content based on language. Text to images \cite{wu2017automatic,reed2016generative}, like DALL-E from OpenAI, can create creative images based on textual descriptions, and significantly improves the efficiency of generating complex paintings. Previously, professional painters had to accumulate materials for years to build complex paintings, but now AI paintings can generate numerous complex paintings in a short period of time. Text-to-video has also yielded satisfactory experimental results \cite{singer2022make,li2018video}. Existing products for text-to-video, such as Lumen5 and CogView2, allow users to input image and text information, such as articles, search queries, or PPTs, to generate videos. However, there is still room for improvement in terms of video duration, clarity, and logic.

In future applications of cross-modal generation, there are several problems that need to be addressed. Firstly, there is a usability issue, where users need to input long text descriptions to obtain high-quality content. This is time-consuming. Secondly, there is a controllability issue. Although text-to-images can generate delicate images quickly, it may not generate images that match specific user requirements. When the model overfits, the image results may not meet expectations. For example, after entering the style description, the model may produce images that do not correspond to the expectations because the style model is overfitting to a specific scene. 

\subsection{Search engine}

Search engines are very suitable for finding websites, but they are often not enough to solve more complex problems or tasks. Every day, there are about 10 billion search queries in the world, but perhaps half of them do not get accurate answers \cite{sekaran2020design}. Now, combined with AIGC technology, it seems that this problem can be changed. With the support of OpenAI technology, Microsoft has updated the Bing search engine and Edge browser. The new version of Bing and Edge integrates search, browsing, and chat into a unified experience. The search engine could provide better search service, more complete answers, a chat experience, and the ability to generate content. Through cooperation with OpenAI, Microsoft has added an advanced AI dialogue model to its search engine. Users can directly communicate with AI chat robots and ask questions in chat interfaces such as ChatGPT. The ChatGPT model could provide fast, accurate, and powerful search capabilities so that it can get the most accurate and relevant answers for basic search queries. In addition, Microsoft has also cooperated with OpenAI to implement special protection measures against harmful content. The Microsoft team is working hard to prevent the propagation of harmful or discriminatory content according to its own principles.

\begin{figure*}[ht]
    \centering
    \includegraphics[clip,scale=0.3]{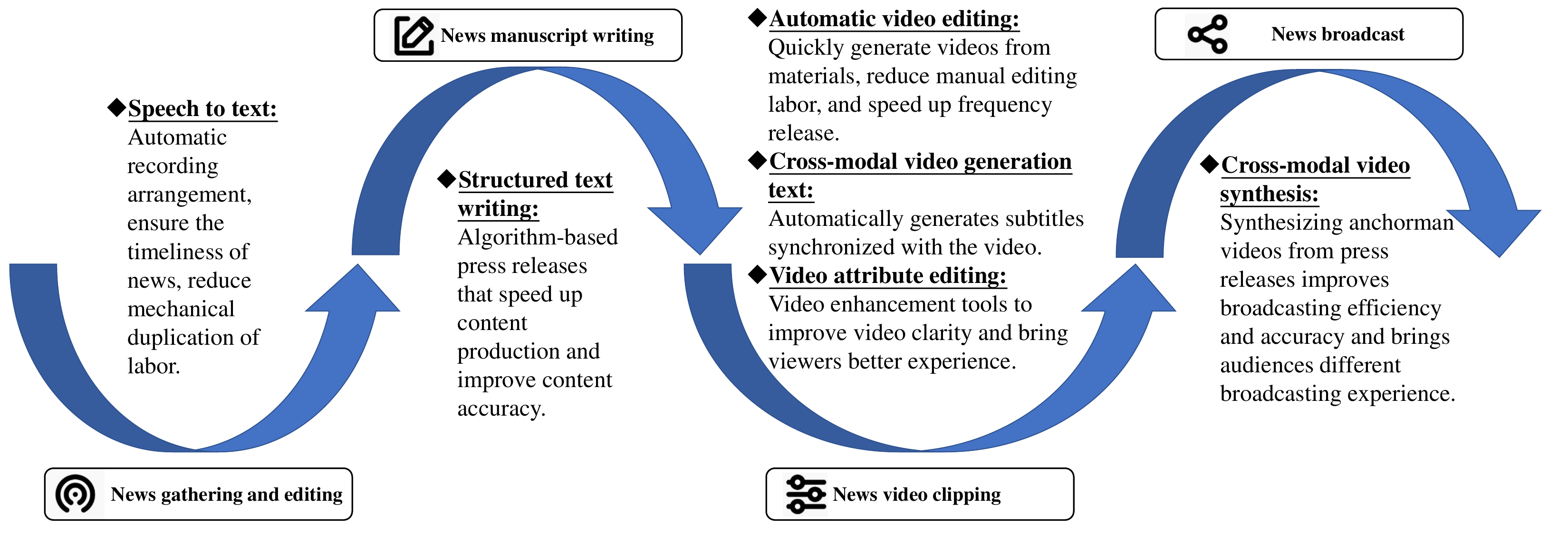}
    \caption{AIGC's empowerment in the field of media.}
    \label{fig:media}
\end{figure*}

\subsection{Media}

AIGC is a game-changer in the media industry. It revolutionizes all aspects of news production, from news collection to manuscript writing, video editing, and news broadcast \cite{de2021artificial,chan2019review}. Fig. \ref{fig:media} illustrates the impact of AIGC on the media industry. By leveraging AIGC, media organizations can improve the efficiency and quality of their content generation and expand their influence after publishing. In news collection, for instance, AIGC can automatically sort and record voice data, which ensures timely news releases. In manuscript writing, the AIGC algorithms combined with structured text writing and press releases can expedite the process of content generation while enabling real-time error correction to enhance accuracy. In video editing, AIGC can perform automatic editing, letter configuration, and video attribute repair. Automatic editing, for example, can significantly reduce manual labor by rapidly generating videos from materials. By leveraging cross-modal generation technology, AIGC can also produce subtitles in sync with the video. Additionally, AIGC's video enhancement tools can improve video clarity. Furthermore, AIGC can synthesize broadcast videos using news text during a news broadcast, which delivers more efficient and accurate results than manual generation.

\subsection{E-commerce}

E-commerce is another mature application field for intelligent text generation. At present, most product titles and descriptions on e-commerce websites, such as JD.com and Taobao \cite{zhang2019automatic,gong2019automatic}, are generated automatically by algorithms. In addition, e-commerce websites commonly implement intelligent customer service systems to address users' inquiries pertaining to shopping, post-sale assistance, and other communication necessities \cite{chen2019antprophet,song2017intension}. The intelligent customer service system must have the ability to accurately comprehend the user's intention and utilize text-generation techniques to generate an appropriate response. Moreover, certain e-commerce websites utilize dialogue summary technology to condense the exchanges between customer service and users into a concise summary \cite{rush2015neural,chopra2016abstractive}. Finally, in order to promote their goods and services, many companies use intelligent text generation technology to generate advertising and marketing copy for their products, which they then disseminate across a variety of multimedia platforms in order to attract users' attention and boost sales \cite{reisenbichler2022frontiers}. It can be seen that intelligent text generation technology has been applied to all aspects of e-commerce, and the use of this technology can reduce the cost of labor.

\subsection{Film}

The combination of AIGC and film has enormous potential to inspire directors with fresh creative ideas \cite{wan2021new}. By assisting with scriptwriting, replacing original roles and settings, and simplifying post-production editing, AIGC can help overcome physical limitations and improve the quality of films. For example, AI technology can analyze vast amounts of script data and generate scripts that fit predetermined styles, which can stimulate directors' creativity. After reviewing and refining the AI-generated script, the director can significantly reduce the time needed for script creation and increase overall productivity. During video capture, AI technology allows for flexible replacement of characters and backgrounds, and can even create digital avatars capable of complex actions. AI can also create virtual scenes and depict scenarios that cannot be captured in real-time. It provides a more immersive viewing experience for audiences. In post-production editing, AI can be used to repair film images and enhance picture quality, as well as quickly generate promotional movie trailers for publicity.

\subsection{Application in other fields}

With big data still in its blooming stage, the growth of AI-powered data-driven technologies will bring more opportunities in the future. In our opinion, AIGC has a wide range of applications beyond the fields mentioned above. For example, in education, AI technology can convert abstract textbooks into concrete visualizations, making it easier for students to learn \cite{lin2022metaverse}. In finance, AI can automatically produce financial information videos and create virtual digital customer service to improve operational efficiency \cite{bahrammirzaee2010comparative}. In healthcare, AI can assist patients in rehabilitation and enhance medical imaging to aid doctors in diagnosing conditions \cite{jiang2017artificial}. Additionally, speech synthesis technology can generate speech audio for individuals with aphasia, enabling them to communicate effectively. In industry, AIGC can rapidly transform digital geometry into real-time 3D models based on physical environments, and digital factories can analyze process flow to reduce design time \cite{ning2021survey}. All in all, there are still too many applications that cannot be listed one-by-one, and need to be further explored.

\section{Conclusion}  \label{sec:conclusion}

With the support of massive amounts of high-quality data and high-performance hardware, a number of algorithms for large models have rapidly developed in recent years. These algorithms possess the ability not only to comprehend text but also to assist in, or automatically generate rich content. Application examples such as ChatGPT have demonstrated the business value and application performance of AIGC technology, leading to widespread attention and investment from numerous front-line companies in a short period of time. This paper provides a brief introduction to AIGC technology and presents its distinct features. Furthermore, we conduct a comparative analysis of the advantages and disadvantages of AIGC capabilities. However, the development of AIGC still faces many challenges and opportunities. We also provide insights into AIGC challenges and future directions. In conclusion, we hope that this review will provide useful ideas for the development of academia, industry, and business, as well as valuable thinking directions and insights for further exploration in the field of AIGC.

\bibliographystyle{IEEEtran}
\bibliography{AIGC}

\end{document}